\documentclass[10pt,twocolumn]{article}

\usepackage[margin=0.72in]{geometry}
\usepackage[T1]{fontenc}
\usepackage{times}
\usepackage{amsmath,amssymb}
\usepackage{graphicx}
\usepackage{booktabs}
\usepackage{microtype}
\usepackage{xcolor}
\usepackage{url}
\usepackage[hidelinks]{hyperref}

\newcommand{\flyvis}{\textit{flyvis}}
\newcommand{\malecns}{\textsc{MaleCNS}}

\usepackage{newtxtext}
\usepackage[varg]{newtxmath}
\usepackage[scaled=0.92]{helvet}
\usepackage{titlesec}
\usepackage[font=small,labelfont={bf,sf,color=accent},textfont=normalfont,
            labelsep=period,skip=5pt]{caption}
\usepackage{enumitem}

\definecolor{accent}{HTML}{0B4F6C}
\definecolor{rulegray}{HTML}{C9C9C2}
\definecolor{panel}{HTML}{F6F6F3}

\titleformat{\section}{\sffamily\bfseries\color{accent}\large}{\thesection}{0.6em}{}
\titleformat{\subsection}{\sffamily\bfseries\color{accent}\normalsize}{\thesubsection}{0.6em}{}
\titleformat{\paragraph}[runin]{\sffamily\bfseries\color{accent}}{}{0em}{}[.]
\titlespacing*{\section}{0pt}{1.1em}{0.45em}
\titlespacing*{\subsection}{0pt}{0.85em}{0.35em}

\title{What Survives on Real Drawings:\\
Active Sampling, Connectome Wiring, and Matched Baselines\\
in Architectural Document Vision}
\author{Dmitry Kuklev\\
Independent Researcher\\
\texttt{dima.kuklev9797@gmail.com}}
\date{September 2026}

\newsavebox{\overviewbox}
\newsavebox{\bandbox}

\begin{document}
\sbox{\overviewbox}{\includegraphics[width=0.985\textwidth]{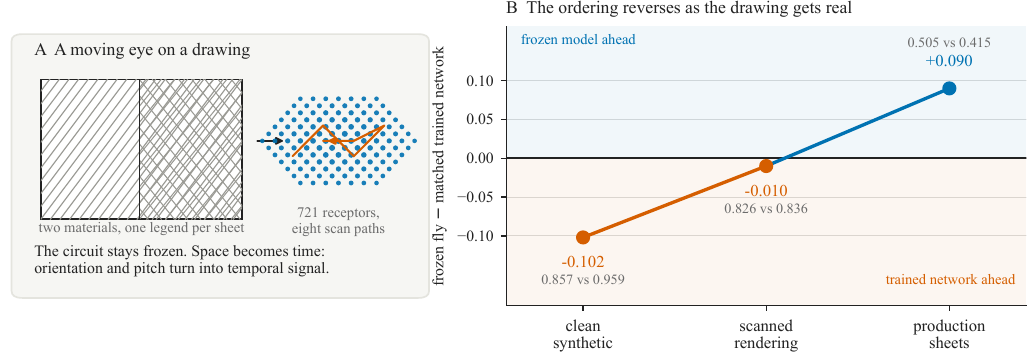}}
\sbox{\bandbox}{\includegraphics[width=0.66\textwidth]{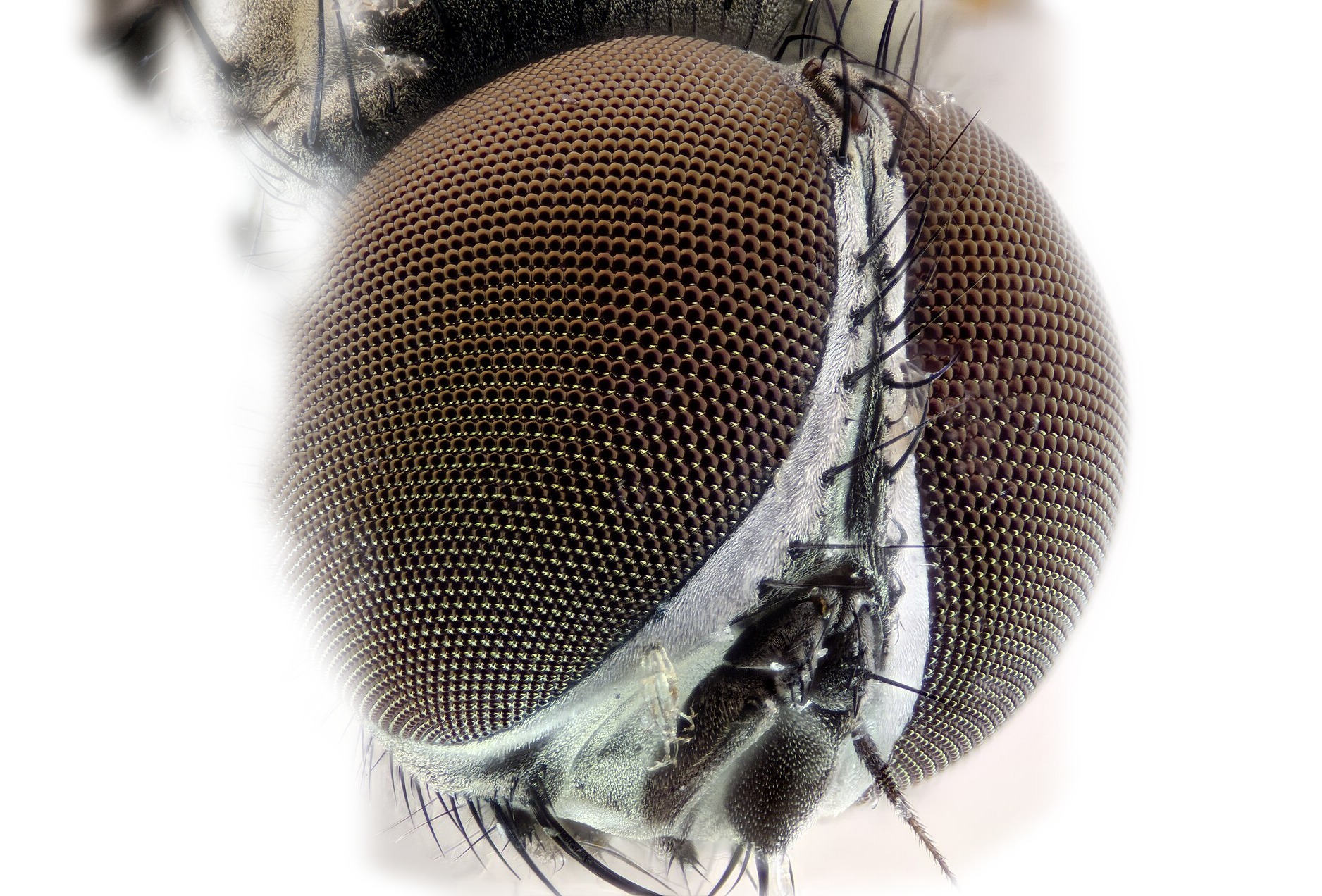}}
\makeatletter
\twocolumn[\begin{@twocolumnfalse}
\noindent\hbox to \textwidth{\hfil\usebox{\bandbox}\hfil}\par
\vspace{-1.2em}
\maketitle
\vspace{-2.2em}
\begin{center}
{\color{rulegray}\rule{0.86\textwidth}{0.5pt}}
\end{center}
\vspace{-0.4em}
\begin{center}
\setlength{\fboxsep}{9pt}\colorbox{panel}{\begin{minipage}{\dimexpr0.86\textwidth-2\fboxsep\relax}
\small\noindent{\sffamily\bfseries\color{accent}Abstract.} A connectome-constrained model of the fly visual system, optimized for motion and then frozen, can be driven over a drawing by prescribed motion and used as a texture representation. We test what that buys on architectural documents while holding the input fixed, so that every baseline sees the same 721 photoreceptor samples. On clean synthetic data the frozen model transfers but loses to task training: 0.857 area-weighted accuracy in one-shot hatch matching against 0.959 for a 5{,}888-parameter network, and 0.619 IoU in wall segmentation against 0.905 for a parameter-matched network that trains $46\times$ faster. That ordering does not survive real documents. Scanning noise and thickened strokes cost the task-trained networks up to 0.188 accuracy and the frozen pipeline 0.030; on a closed set of fourteen production sheets, opened once, a 1{,}876-parameter fly model reaches 0.505 average precision against 0.415 for a network two hundred times larger, and on sheets annotated by a quantity surveyor the convolutional model effectively abstains. A held-out split, opened once under preregistration, reproduces that ordering: 0.835 for the frozen circuit, 0.894 for a receptors-only descriptor built from the same moving signal, and 0.971 to 0.980 for the trained networks. The wiring itself is load-bearing. On the development split, rewiring the connectome while preserving degrees, and separately while preserving type pairs and transmitter signs, costs 0.271 to 0.356 accuracy across three seeds, and a positive control confirms the test can see a known structural effect. What the intact circuit computes is nevertheless no more useful than its own retina: receptors lead it by 0.059 on the held-out split, the circuit adds nothing in front of a convolutional decoder, and silencing the motion-output types T4 and T5 leaves both tasks intact. Exact structure remains insufficient to beat what the moving retina already provides.

\vspace{0.35em}\noindent {\sffamily\bfseries\color{accent}Keywords.} connectome-constrained networks; active vision; domain shift; architectural drawings; matched null models
\end{minipage}}
\end{center}
\vspace{0.2em}
\end{@twocolumnfalse}]
\makeatother

\section{Introduction}
\begin{figure*}[t]
  \centering
  \includegraphics[width=0.985\textwidth]{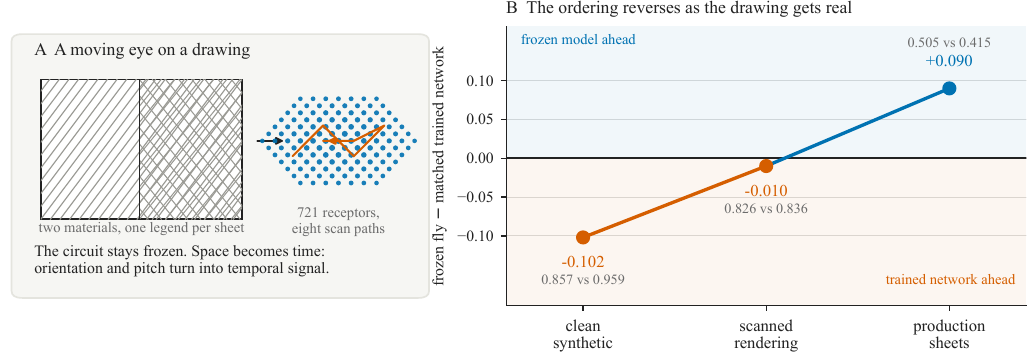}
  \caption{The study at a glance. \textbf{A}~A frozen fly visual model is driven over a drawing patch by
  prescribed motion, so that hatch orientation and pitch become a temporal signal on 721 receptors.
  \textbf{B}~The ordering of models is not a property of the models: each point is the margin of the frozen
  pipeline over a trained network on identical input, in that regime's own metric. Task training wins by 0.102
  on clean synthetic data, scanning all but removes the margin, and on a closed set of fourteen production
  sheets, opened once, the 1{,}876-parameter frozen model is 0.090 average precision ahead of a convolutional
  network with two hundred times more parameters. Absolute scores are low for every model on real sheets
  (Table~\ref{tab:exam}); what reverses is the ranking.}
  \label{fig:overview}
\end{figure*}
Anatomy can now be turned into a running model. Networks constrained by the fly optic-lobe connectome predict measured activity across cell types~\cite{lappalainen2024,turagalab}, and whole-brain graph models drive body control~\cite{jin2026}. The engineering reading of this is tempting: a circuit with a few hundred free parameters, shaped by evolution, reused as a visual front end for tasks it never saw.

Architectural drawings are a good place to check that reading, for three reasons. Their content is exactly what a texture representation should capture, since hatch patterns differ in orientation and pitch while sharing mean brightness. Exact ground truth is available from a renderer. And the comparison can be made fair in a way that is rare in vision: a fly model and a convolutional network can be restricted to the same 721 photoreceptor samples, so that a difference in score is a difference in computation, not in input.

We ran two benchmarks under that discipline. The first is one-shot hatch matching: given a legend strip per material on the sheet, classify wall regions by texture. The second is wall segmentation from a rendered plan. Both were also carried out of the laboratory and onto real production documents, where drawings are scanned, dense with tables and dimension chains, and annotated only in part.

The results separate into a clean-data story and a real-document story, and they disagree (Figure~\ref{fig:overview}). On clean data, task-trained networks win everywhere, by margins that no framing removes. On shifted renderings and on real sheets, the ordering flips or collapses into instability. The mechanistic analysis explains why the frozen pipeline is the more stable of the two, and separates two things that are easy to conflate. The exact wiring is load-bearing: every matched rewiring we could build substantially degrades the result. It is also not where the usable margin comes from, since the moving retina in front of the circuit already carries it.

\paragraph{Contributions}
Two independent benchmarks evaluated under strict information matching; preregistered gates with cluster bootstrap intervals on one, a preregistered three-seed topology and temporal protocol with a hash-chained run registry on the other; three convergent controls that separate active sampling from wiring; a parameter-matched negative result on a test set opened once, with compute accounting; and a real-document evaluation, including a closed production set, that identifies where each family of models actually fails.

\section{Related Work}
Lappalainen et al.~\cite{lappalainen2024} constrain a recurrent network by the optic-lobe connectome and fit a small set of biophysical parameters, releasing trained individuals through \flyvis{}~\cite{turagalab}. Whole-brain connectomic models have been applied to locomotion~\cite{jin2026}, and the \malecns{} release provides a complete male central nervous system connectome with predicted transmitters~\cite{malecns}. Work on matched nulls is the reason our topology claim is stated cautiously: apparent structural advantages often shrink once degree- and type-preserving rewirings are used as controls~\cite{dhiman2026}.

Micro-saccadic sampling converts fine spatial structure into a temporal signal in the fly~\cite{juusola2017}; in computer vision the underlying idea is classical active vision~\cite{aloimonos1988}. Our stimulus is deliberately simple: the raster is translated along prescribed paths, which is an engineering device rather than a model of behaviour.

For texture, Gabor filters~\cite{gabor1946} and local binary patterns~\cite{ojala2002} remain the reference hand-designed descriptors, and frozen large encoders~\cite{oquab2023,tschannen2025} the reference learned ones. Floor-plan geometry comes from CubiCasa5K~\cite{kalervo2019} and a public annotated corpus~\cite{cis2025}.

\section{Benchmarks and Data}
\subsection{Hatch matching}
Each sheet provides wall regions, one legend strip per material class present on that sheet, and the raster. Classes are local: the model matches query texture to legend texture and never needs the name of a material. Four $64\times64$ windows are taken from each legend strip and three from each of up to forty regions. The score is area-weighted accuracy,
\begin{equation}
A_s=\frac{\sum_j a_{sj}\,\mathbf{1}[\hat y_{sj}=y_{sj}]}{\sum_j a_{sj}},\qquad A=\frac{1}{N}\sum_s A_s .
\end{equation}
The generator combines CubiCasa5K geometry with procedural hatches. The development split has 400 sheets in 257 source-plan clusters and four conditions of 100 sheets: orientation varies while simple intensity statistics are made uninformative; pitch varies with proportional stroke width; and two conditions vary pattern similarity. Intervals are cluster bootstraps over source plans. The held-out test split was opened exactly once, under a preregistration whose hash was fixed beforehand, with the comparison set, the verdict rule and the abstention thresholds all frozen on development data. It was not reopened.
One-shot here means one physical legend strip per class on the sheet being read, sampled through four windows; no material class is ever seen during fitting. Standardization is per sheet and uses the unlabelled query descriptors of that sheet, so the protocol is transductive within a sheet and does not model a single-window decision in isolation.

\paragraph{The benchmark is the second version of itself} The first version had three defects. Nineteen per cent of development sheets contained at least one pair of material classes that rendered indistinguishably. Material identity correlated with wall thickness and piece size. And a second standardization inside the evaluation path could cancel the feature weights it was meant to apply. Those defects flattered every biological reading we had at the time. The rebuild checks every class pair for rendered distinguishability and leaves none failing, assigns materials independently of geometry, standardizes once, resamples bootstrap clusters by source plan, seals the test split behind a guard, and routes every measurement through a single evaluation library that passes sixteen unit tests. The comparisons in Section~\ref{sec:clean} were registered only afterwards. We describe the repair because it changed conclusions and not merely numbers: a benchmark defect is a reliable way to manufacture a convincing biological result.

\subsection{Wall segmentation}
An exact renderer produces wall masks without annotation noise. The manifest holds 11{,}238 fitting sources, 64 for model selection, 256 for calibration and 256 for a final test that was opened once, with the decision threshold fixed on calibration beforehand. Models emit one logit per readout column; we report average precision, ROC-AUC and IoU.

\subsection{Real documents}
Three kinds of real data are used. A public corpus of 494 annotated plans~\cite{cis2025} supports direct training and validation. Sixty sheets from sixteen projects carry vector CAD layers, which give dense and reliable masks. Finally, a closed set of fourteen production sheets from five projects, whose masks come from native PDF fills, was opened once as an examination; because fills omit line-only partitions, every model is penalized for finding them. A separate family of sheets carries quantity-surveyor takeoffs, which are partial by construction: unmarked regions are unknown rather than negative, so those sheets support recall-style statements only.

\section{Models}
\begin{figure*}[t]
  \centering
  \includegraphics[width=0.96\textwidth]{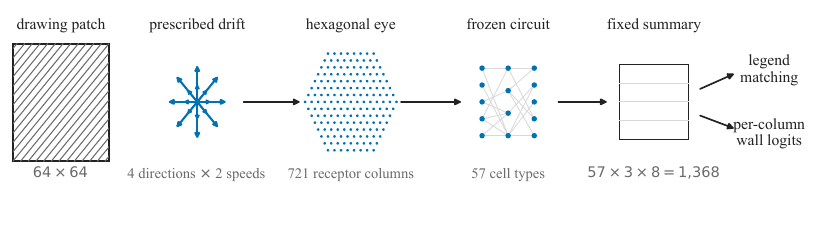}
  \caption{The pipeline. A drawing patch is translated along eight prescribed trajectories, rendered through the fly model's own hexagonal eye, passed through the frozen connectome-constrained circuit, and summarized by three statistics per cell type. The same representation feeds one-shot legend matching and per-column wall prediction. No hatch or wall label updates the circuit.}
  \label{fig:pipeline}
\end{figure*}

\subsection{Frozen fly descriptor}
A $64\times64$ window (Figure~\ref{fig:pipeline}) is resized to the model eye and translated along four directions at two speeds, giving eight trajectories of twelve frames at a 20\,ms step, of which the first four are settling. For each of 57 non-input cell types we record mean activity, temporal standard deviation and spatial variation, which yields a descriptor of $57\times3\times8=1{,}368$ dimensions. Legend and query descriptors are standardized per sheet, averaged, L2-normalized and compared by cosine similarity. No hatch label ever updates the model.

\subsection{Trained connectome readouts}
For segmentation, two connectome models are used. The \flyvis{} optic lobe with a shared spatial type readout has 1{,}876 trainable parameters over the same 721 receptors. The whole-brain \malecns{} readout reduces the connectome to a signed, contact-weighted operator over 166{,}700 cells and 6{,}063{,}706 edges, with 23{,}523 trainable parameters covering per-type gains, time constants and readout weights. In both, the wiring is frozen and only physiology is fitted.

\subsection{Matched competitors}
Every competitor receives the fly model's own input. For hatch matching these are eleven pixel statistics, Gabor, local binary patterns, five frozen large-encoder checkpoints under prescribed pooling recipes, a receptors-only descriptor built from the same moving photoreceptor signal without the connectome, and small task-trained networks of 5{,}888 to 92{,}672 parameters. For segmentation they are a pixel-darkness rule, linear and perceptron readouts of the same samples, and convolutional networks on the receptor lattice from 1{,}934 to 365{,}921 parameters. A network that sees the full crop is reported only as a ceiling.

\subsection{Evidence levels}
Results are labelled. \textbf{P} means preregistered, with comparison and verdict rule fixed before the numbers existed. \textbf{A} adds an audit: preregistration hash, hash-chained run registry, verified sidecars. \textbf{R} means recomputed from saved per-item values during manuscript preparation. \textbf{L} is a logged project record with fixed seed and configuration. \textbf{E} is exploratory, single seed. Every number in this paper maps to a stored record listed in the accompanying ledger.

\section{Clean Data}
\label{sec:clean}
\subsection{The frozen model transfers}
\begin{table}[t]
\centering
\caption{Hatch matching, raw legend, 400 development sheets.}
\label{tab:main}
\small
\setlength{\tabcolsep}{3pt}
\begin{tabular}{lrrrrr}
\toprule
Representation & All & Angle & Pitch & Hard & Easy\\
\midrule
Uniform guess & 0.321 & 0.315 & 0.378 & 0.277 & 0.314\\
Fly + drift (P,R) & 0.857 & 0.840 & 0.793 & 0.848 & 0.945\\
\quad extended legend & 0.910 & 0.975 & 0.827 & 0.875 & 0.961\\
\quad random init (L) & 0.662 & -- & -- & -- & --\\
Receptors only (P,R) & 0.891 & 0.967 & 0.790 & 0.853 & 0.956\\
Gabor only (P,R) & 0.802 & 0.900 & 0.714 & 0.742 & 0.850\\
LBP only (P,R) & 0.702 & 0.373 & 0.823 & 0.707 & 0.908\\
11 pixel statistics (P,R) & 0.691 & 0.299 & 0.675 & 0.858 & 0.933\\
DINOv2-B (L) & 0.757 & 0.581 & 0.714 & 0.793 & 0.939\\
DINOv2-S (L) & 0.785 & 0.682 & 0.738 & 0.781 & 0.939\\
DINOv3-S (L) & 0.767 & 0.640 & 0.736 & 0.768 & 0.925\\
PE-core B (L) & 0.792 & 0.627 & 0.766 & 0.817 & 0.957\\
SigLIP 2 B (L) & 0.804 & 0.656 & 0.785 & 0.819 & 0.955\\
\midrule
CNN 5.9k, untrained (R) & 0.751 & 0.818 & 0.690 & -- & --\\
CNN 5.9k, trained (R) & \textbf{0.959} & -- & -- & -- & --\\
\bottomrule
\end{tabular}
\end{table}

\begin{figure}[t]
  \centering
  \includegraphics[width=\columnwidth]{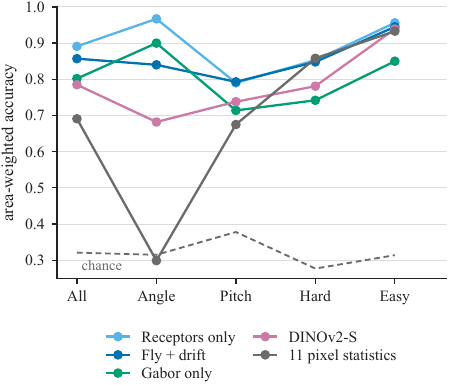}
  \caption{Hatch matching by controlled condition, 400 development sheets. On the orientation condition the benchmark was validated in advance to make intensity statistics uninformative, and the pixel baseline falls to chance while the fly descriptor does not.}
  \label{fig:conditions}
\end{figure}

The frozen pipeline reaches 0.857 over 400 sheets (Table~\ref{tab:main}), and 0.910 when the reference legend shows more of the repeating pattern, a paired gain of $+0.053$ $[0.039, 0.067]$. The orientation condition is the sharpest test, because the benchmark was validated in advance to make simple intensity statistics useless there: eleven pixel statistics score $-0.016$ $[-0.051, 0.020]$ relative to a uniform guess. The fly descriptor reaches 0.840 on that condition against 0.299, a paired difference of $+0.541$ $[0.490, 0.590]$. Figure~\ref{fig:conditions} shows the pattern by condition. It is also ahead of every large frozen encoder we tested under its prescribed pooling recipe: by $+0.100$ $[0.077, 0.123]$ over DINOv2-B, $+0.072$ $[0.051, 0.091]$ over DINOv2-S, $+0.089$ $[0.069, 0.110]$ over DINOv3-S, $+0.065$ $[0.044, 0.086]$ over PE-core B and $+0.053$ $[0.033, 0.072]$ over SigLIP~2~B. A frozen circuit that has never been shown a drawing reads these textures better than encoders pretrained on hundreds of millions of natural images.

The transfer depends on the motion pretraining and not merely on the architecture. Re-initializing the same wiring at random drops the score to 0.662, a gap of $+0.195$ $[0.174, 0.216]$ in favour of the pretrained parameters. All results here use pretrained individual~001; three further individuals of the same ensemble score 0.819, 0.823 and 0.854, so the choice of individual matters as much as several of our design decisions do.

\subsection{Moving the input matters}
Motion is what turns a spatial pattern into a temporal one. For a periodic hatch
\begin{equation}
I(\mathbf{x})=a+b\cos(2\pi\,\mathbf{k}\cdot\mathbf{x}+\phi)
\end{equation}
translated at velocity $\mathbf{v}$, the temporal frequency reaching a photoreceptor is
\begin{equation}
f_t\propto\lvert\mathbf{k}\cdot\mathbf{v}\rvert ,
\end{equation}
so hatch orientation and pitch both become frequencies, and a single direction cannot separate them: sweeping four directions at two speeds resolves the ambiguity. A circuit optimized for motion prediction is, by construction, in a position to read those frequencies, which is the reason to expect transfer at all.

On the 200 sheets of the two equalized conditions, replacing drift with a repeated centre frame lowers accuracy from 0.816 to 0.727, a paired difference of $+0.089$ $[0.066, 0.112]$. Enlarging the modelled eye from 721 to 1{,}951 columns at the same speed in paper space adds $+0.031$ $[0.022, 0.041]$.

Segmentation reproduces the effect under a stricter protocol. With three seeds, a preregistration hash and a verified run registry, replacing the four-step return trajectory by a repeated static frame costs 0.062 IoU and 0.050 average precision on clean data, and 0.056 IoU and 0.123 average precision under procedural clutter. The hierarchical paired difference on clean IoU is $0.074$ $[0.055, 0.094]$, positive in every seed and every preregistered measure. The gain is largest under clutter, which is what an added temporal channel should do.

A third and weaker record agrees. On the same segmentation task, letting the model take more looks at a sheet instead of one raised average precision from 0.900 with a single glimpse to 0.913 with nine and 0.955 with twenty-five (L). Sampling helps in all three settings we can measure it in, with effects an order of magnitude apart in size.

\begin{figure*}[t]
  \centering
  \includegraphics[width=0.96\textwidth]{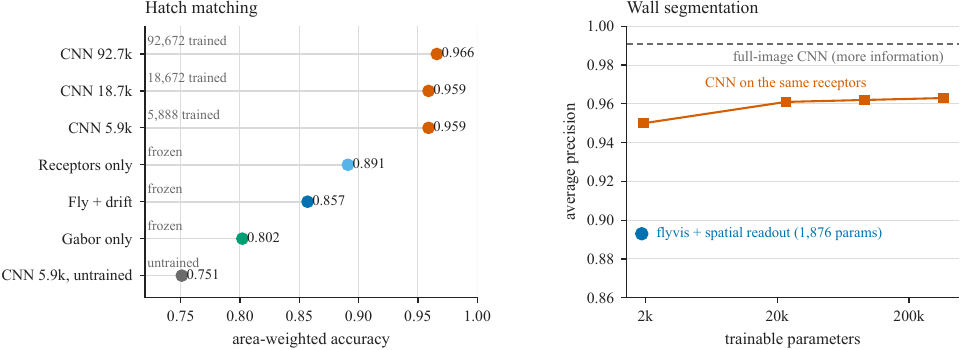}
  \caption{Matched comparisons on clean data. Left: hatch matching, frozen descriptors against task-trained networks. Right: wall segmentation on identical receptor input, where a convolutional network with roughly the same parameter count as the fly readout is already ahead and larger ones saturate, indicating that the ceiling is set by the eye rather than by the decoder.}
  \label{fig:capacity}
\end{figure*}

\subsection{Task training wins on clean data}
\begin{table}[t]
\centering
\caption{Matched comparisons on clean data. The two rows marked (A) come from the segmentation test set opened once; the others are development results.}
\label{tab:matched}
\small
\setlength{\tabcolsep}{3pt}
\begin{tabular}{llrr}
\toprule
Comparison & Params & Fly & Matched\\
\midrule
Hatch, task CNN (R) & 5{,}888 & 0.857 & \textbf{0.959}\\
Hatch, Gabor, angle (P,R) & -- & 0.840 & \textbf{0.900}\\
Walls, small CNN (L) & ${\approx}1.9$k & 0.893 & \textbf{0.950}\\
Walls, IoU (A) & ${\approx}23$k & 0.619 & \textbf{0.905}\\
Walls, AP (A) & ${\approx}23$k & 0.730 & \textbf{0.984}\\
\bottomrule
\end{tabular}
\end{table}

All nine recorded task-trained runs on hatch matching (Table~\ref{tab:matched}), three sizes by three seeds, exceed the frozen descriptor, with means of 0.959, 0.959 and 0.966. The same networks before training score 0.751 and 0.743, below the frozen descriptor by $+0.106$ $[0.086, 0.125]$ and $+0.114$ $[0.094, 0.133]$. Both sides of this comparison are therefore trained representations. One was fitted to this corpus; the other was fitted to motion prediction in a different domain and then frozen. In segmentation the preregistered pilot already favoured the network: 0.719 against 0.545 IoU on clean data, 0.477 against 0.336 under clutter. The final test agrees. A 23{,}225-parameter convolutional network reaches 0.905 IoU and 0.984 average precision, where the parameter-matched 23{,}523-parameter whole-brain readout reaches 0.619 and 0.730. It wins on all 256 test sources, with a paired macro-IoU difference of $-0.287$ $[-0.299, -0.275]$, and it trains in 166\,s within 28\,MiB against 7{,}622\,s and 778\,MiB. A specialized Gabor bank likewise beats the fly descriptor on orientation, by $0.060$ $[0.024, 0.097]$, although the fly is ahead on pitch by $0.079$ $[0.047, 0.112]$.

Figure~\ref{fig:capacity} places these comparisons on a parameter axis. On clean, in-distribution data there is therefore no case for the biological pipeline as a practical choice.

\subsection{The held-out split}
\begin{table}[t]
\centering
\caption{Held-out hatch split, opened once under preregistration. Coverage is the share of decisions accepted at a confidence threshold fixed on development data; accepted accuracy is measured on that share.}
\label{tab:test}
\small
\setlength{\tabcolsep}{3pt}
\begin{tabular}{lrrrr}
\toprule
Model & Accuracy & ECE & Coverage & Accepted\\
\midrule
Fly + drift (P,A,R) & 0.835 & 0.017 & 0.755 & 0.913\\
Receptors only (P,A,R) & 0.894 & 0.048 & 0.887 & 0.921\\
Gabor only (P,A,R) & 0.803 & 0.010 & 0.579 & 0.925\\
DINOv2-S (P,A,R) & 0.779 & 0.034 & 0.406 & 0.905\\
CNN 5.9k (P,A,R) & 0.971 & 0.012 & 1.000 & 0.971\\
CNN 18.7k (P,A,R) & \textbf{0.980} & 0.010 & 1.000 & 0.980\\
CNN 92.7k (P,A,R) & 0.976 & 0.012 & 1.000 & 0.976\\
\bottomrule
\end{tabular}
\end{table}

The split was opened once, by a single process, with the comparison set and the verdict rule hashed in advance and no model trained or tuned on it. Table~\ref{tab:test} shows that the development ordering survives: receptors lead the circuit by $+0.059$ $[0.043, 0.075]$, and the trained networks lead it by $+0.136$ to $+0.144$. Nothing about the clean-data conclusion changes when the data are new.

Allowing a model to decline the cases it is unsure of changes the picture only in degree. At a confidence threshold fixed on development data, the frozen circuit answers 75.5 per cent of regions at 0.913 accuracy and the receptors-only descriptor 88.7 per cent at 0.921, while the trained networks clear the same bar without declining anything. The circuit's threshold does not quite transfer: its accepted accuracy on the held-out split is 0.913 $[0.897, 0.927]$, whose lower bound sits just under the 0.90 guarantee the threshold was chosen to give, whereas the receptors-only descriptor keeps it at 0.921 $[0.906, 0.936]$. Abstention makes a frozen descriptor usable where it is confident; it does not close the gap to a trained network.

\section{Shift and Real Documents}
\subsection{Rendering shift}
\begin{table}[t]
\centering
\caption{Hatch matching under rendering shift (E, single seed for the trained networks). Values are area-weighted accuracy.}
\label{tab:shift}
\small
\setlength{\tabcolsep}{3pt}
\begin{tabular}{lrrrr}
\toprule
Representation & Clean & Scan & $\Delta$ scan & $\Delta$ thick\\
\midrule
Fly + drift & 0.857 & 0.826 & $-0.030$ & $-0.068$\\
Receptors only & 0.891 & 0.888 & $-0.003$ & $-0.066$\\
Gabor + pixel & 0.841 & 0.845 & $+0.004$ & $-0.055$\\
DINOv2-S & 0.785 & 0.706 & $-0.078$ & $-0.022$\\
CNN 5.9k & 0.966 & 0.887 & $-0.080$ & $-0.168$\\
CNN 18.7k & 0.966 & 0.778 & $-0.188$ & $-0.176$\\
CNN 92.7k & 0.968 & 0.843 & $-0.125$ & $-0.176$\\
\bottomrule
\end{tabular}
\end{table}

\begin{figure}[t]
  \centering
  \includegraphics[width=\columnwidth]{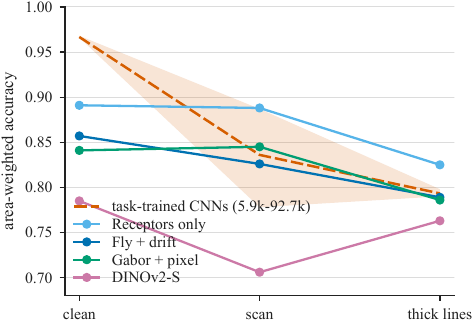}
  \caption{Rendering shift. Task-trained networks hold the highest clean scores and lose the most under scanning noise and thickened strokes; the frozen sampling descriptors change least. Single seed for the trained networks.}
  \label{fig:shift}
\end{figure}

Scanning noise and thickened strokes are the two most common ways a drawing arrives in practice not looking like its source; Table~\ref{tab:shift} gives the whole matrix. They cost the task-trained networks between 0.080 and 0.188 accuracy under scan (Figure~\ref{fig:shift}) and about 0.17 under thickening, while the frozen sampling descriptors lose 0.030 and 0.003 under scan. The clean-data margin of roughly 0.10 in favour of task training narrows to near parity for the smallest network and reverses for the largest. Nothing in the frozen pipeline was fitted to the rendering statistics of the benchmark, which is the straightforward explanation.

\subsection{Production sheets}
\begin{table}[t]
\centering
\caption{Closed production examination: fourteen sheets, five projects, opened once. Masks come from native PDF fills, so line-only partitions are absent from the reference.}
\label{tab:exam}
\small
\setlength{\tabcolsep}{3pt}
\begin{tabular}{lrrr}
\toprule
Model & Params & AP & IoU\\
\midrule
Fly, colour eye, synapse gains & 1{,}560{,}776 & \textbf{0.539} & \textbf{0.407}\\
Fly, colour eye & 1{,}876 & 0.505 & 0.366\\
CNN, same channels & 367{,}073 & 0.415 & 0.281\\
CNN, grayscale & 365{,}921 & 0.346 & 0.277\\
\bottomrule
\end{tabular}
\end{table}

\begin{figure*}[t]
  \centering
  \includegraphics[width=0.96\textwidth]{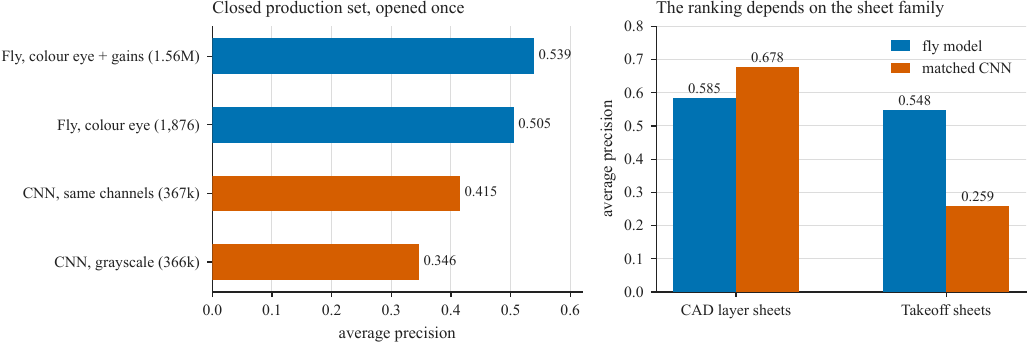}
  \caption{Real documents. Left: the closed production examination, opened once, where the smaller fly model leads convolutional networks with two hundred times more parameters on the same channels. Right: the same two families on two sheet types, showing that the ranking follows the sheet family rather than the model.}
  \label{fig:real}
\end{figure*}

On the closed examination (Figure~\ref{fig:real}) the ordering of the clean benchmarks is reversed: the smaller fly model, with 1{,}876 parameters, is 0.090 average precision ahead of a convolutional network with two hundred times more parameters and the same input channels. Absolute quality is low for every model, with the best IoU at 0.41, and part of the gap is an artefact of the reference, since fill-derived masks contain no line-only partitions and penalize whichever model finds them. The comparison is nevertheless matched and was run once, and it is the only real-document comparison in this work that was fixed in advance.

The behaviour behind the numbers is visible without any metric. On sheets annotated by a quantity surveyor, the convolutional model effectively abstains: it marks eight to thirty times less area than the fly model, in eighteen of twenty sparse sheets and in all three dense ones. On whole dense takeoff sheets the fly model reaches 0.88 average precision against 0.37, although with three sheets this is an observation rather than an estimate. On clean vector CAD layers the situation is the opposite and the convolutional model leads, 0.678 against 0.585 average precision.

\begin{figure*}[tb]
  \centering
  \includegraphics[width=0.94\textwidth]{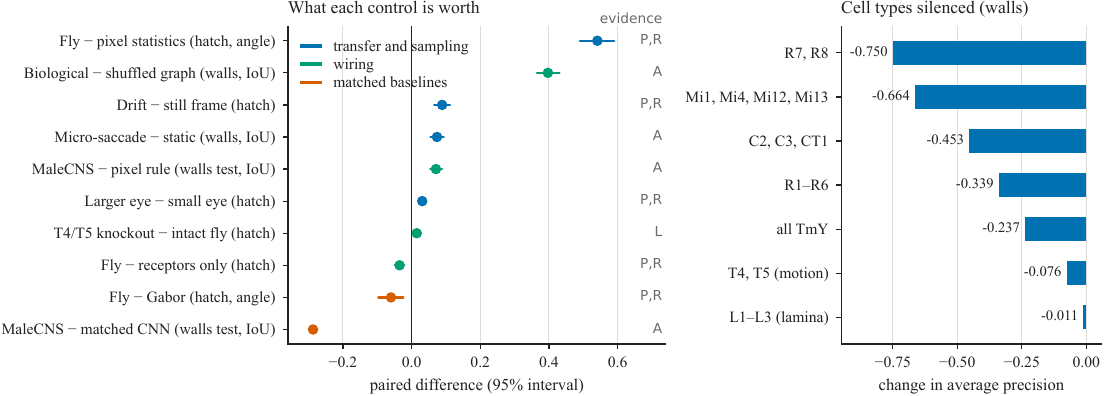}
  \caption{Left: every control with its paired difference and 95\% interval, coloured by what it tests. Right: cell types silenced in the trained segmentation model; the motion detectors T4 and T5 are nearly dispensable while the input stage and the medulla cascade are not.}
  \label{fig:effects}
\end{figure*}

\begin{figure*}[tb]
  \centering
  \includegraphics[width=0.94\textwidth]{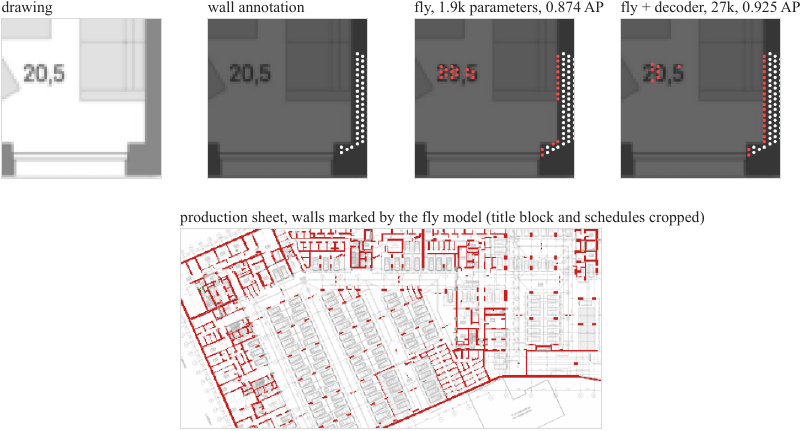}
  \caption{Top: one plan from the public corpus. White dots are the wall annotation and are repeated unchanged in every panel as a reference; red dots are columns the model marked as wall that lie outside that annotation. Both fly-based models make the same kind of error, an extra column along the inner edge of the wall, and both also fire on the dimension text. Scores are corpus-level average precision on this public set, where every model is trained directly: 0.874 AP for the fly readout and 0.925 AP once a convolutional decoder is placed behind it. A convolutional network on the same receptors and without the circuit reaches 0.919 there, so the fly front end adds little to a trained decoder, as Section~\ref{sec:mechanism} reports. Bottom: a production sheet with walls marked by the fly model, cropped to the plan area; the title block and schedules are removed. That is the regime in which the ordering reverses (Table~\ref{tab:exam}).}
  \label{fig:qualitative}
\end{figure*}

\subsection{Rankings are not stable}
Three further observations temper any practical claim.

Figure~\ref{fig:qualitative} shows what the outputs look like on a public plan and on a production sheet. First, model ranking depends on which annotation authority is treated as truth. On the same real sheets, fill-derived masks put the fly model ahead, 0.539 against 0.415, while full CAD-layer masks reverse it, 0.778 against 0.844.

Second, ranking depends on the training mixture. Tuned towards the takeoff sheets, a fly configuration reaches 0.548 average precision there against 0.259 for the convolutional model; tuned differently, the same family scores 0.325 where the convolutional model reaches 0.48 to 0.50. Specialists beat generalists on each domain, and a fixed blend is worse than either.

Third, transfer without adaptation does not work for anyone. Zero-shot from synthetic renders to real plans gives 0.224 average precision for the fly model, 0.504 for a matched perceptron and 0.283 for a full-image network. Real data must be trained on directly, and the practically useful mechanism turned out to be selecting a specialist per sheet, using the observation that the convolutional model falls silent on raster-like sheets.

\section{Where the Advantage Comes From}
\label{sec:mechanism}
\subsection{The wiring matters to the model}
Both benchmarks now have a matched graph null, and both answer the same way.

On hatch matching, and on the development split, since the held-out split was opened only for the model comparison, rewiring the optic-lobe connectome while holding the edge count at 1{,}513{,}231 over 45{,}669 nodes drops accuracy from 0.857 to 0.539 when in- and out-degrees are preserved, to 0.586 when type pairs and transmitter signs are preserved as well, and to 0.500 for an Erd\H{o}s--R\'enyi graph of the same density: paired differences of $+0.318$ $[0.296, 0.339]$, $+0.271$ $[0.248, 0.293]$ and $+0.356$ $[0.334, 0.379]$, each over three rewiring seeds with the invariants verified and the wiring hashed. The design carries its own positive control: silencing the R7 and R8 input types, a structural change with a known effect, costs $0.092$ $[0.079, 0.104]$, so the protocol can detect a real structural manipulation rather than merely rewarding any intact graph.

On segmentation, a null graph that preserves edge count, per-node in- and out-degree, incoming contact sums, type-pair counts and transmitter signs, while destroying the exact wiring, collapses segmentation across three seeds: from 0.483 to 0.089 IoU, and from 0.964 to 0.514 ROC-AUC. The hierarchical paired difference is $0.397$ $[0.365, 0.431]$. Exact wiring is clearly load-bearing for that model. On the fixed test opened once, the same readout also beats a pixel-darkness rule, 0.619 against 0.552 IoU, a paired macro-IoU difference of $+0.071$ $[0.054, 0.089]$. The pixel rule keeps the higher average precision, so the model wins at its operating threshold rather than in global ranking quality.

\subsection{But the useful computation is in front of it}
\begin{table}[t]
\centering
\caption{Controls that separate sampling from wiring.}
\label{tab:controls}
\small
\setlength{\tabcolsep}{3pt}
\begin{tabular}{llr}
\toprule
Control & Task & Effect\\
\midrule
Drift $-$ still frame (P,R) & Hatch & $+0.089$ $[0.066, 0.112]$\\
Micro-saccade $-$ static (A) & Walls & $+0.074$ $[0.055, 0.094]$\\
Biological $-$ shuffled (A) & Walls & $+0.397$ $[0.365, 0.431]$\\
Fly $-$ receptors only (P,R) & Hatch & $-0.035$ $[-0.049, -0.021]$\\
Fly front end, same decoder (L) & Walls & $+0.001$\\
T4/T5 knockout $-$ fly (L) & Hatch & $+0.015$ $[0.010, 0.021]$\\
Random 8-type knockout (L) & Hatch & $-0.046$ to $-0.070$\\
T4/T5 knockout (L) & Walls & $-0.076$\\
R7/R8 knockout (L) & Walls & $-0.750$\\
\bottomrule
\end{tabular}
\end{table}

Three controls, listed in Table~\ref{tab:controls} and summarized with every other comparison in Figure~\ref{fig:effects}, point the same way. A descriptor built from the same moving photoreceptor signal, with no connectome behind it, scores 0.891 against the network's 0.857. Putting the optic-lobe model in front of a convolutional decoder yields 0.962 average precision where the same decoder on raw receptors yields 0.961, so the circuit passes the available information through without adding to it. And silencing the motion-output types T4 and T5, the very pathway the model was optimized for, leaves hatch matching marginally better than intact and costs segmentation only 0.076, while silencing eight random types costs 0.046 to 0.070 and silencing R7/R8 costs 0.750. The representation is distributed and structure-dependent, but the structure that carries the task is the input stage and the medulla cascade behind it, not the motion detectors.

Two attempts to buy an advantage back inside the model failed in the same direction. Fine-tuning 734 of the fly model's own parameters on the synthetic training split lifts hatch accuracy from 0.809 to 0.892 at best -- ordinary gradient descent, not biological plasticity, and it arrives only where the untrained receptors-only descriptor already sits. A mushroom-body-inspired readout gave no reliable gain either. It expands the descriptor into two thousand Kenyon-like units, each drawing on seven inputs, under a five-per-cent winner rule. Against a plain linear head it scores 0.874 against 0.880 on the fly descriptor, 0.896 against 0.888 on Gabor, and 0.880 against 0.890 on DINOv2-S. Adding more fly anatomy to the readout did not add accuracy.

\subsection{The stimulus never completes a cycle}
Arithmetic gives a sharper reading of the two negative results above. The model eye renders a $64\times64$ window into a frame of 391 pixels, a magnification of $6.1\times$, with receptors spaced 13 frame pixels apart. The two prescribed speeds are 3 and 6 frame pixels per step, that is 0.49 and 0.98 window pixels, so over the eight frames that enter the readout the pattern travels between 3.9 and 7.9 window pixels. The hatch period in this benchmark is 8.6 to 22.9 window pixels for the fine patterns and larger for the coarse ones. The stimulus therefore delivers between roughly a fifth and nine tenths of one spatial period, and never a full cycle.

By equation~(3) the temporal frequency is $1.6$ to $4.9$\,Hz. That lies inside the range the motion pathway is tuned to. But the observation window is 0.16\,s, shorter than one cycle of it. A frequency cannot be estimated from less than a period. Under this stimulus the motion-output types have no completed temporal frequency to encode, so finding T4 and T5 dispensable is what the design predicts, and it is weak evidence about the circuit. What drift does supply here is a small set of sub-period phase shifts, which is enough to break the ambiguity between orientations but not enough to read pitch as a frequency. This is also the most likely reason the frozen circuit adds nothing in front of a convolutional decoder: the quantity it is built to compute is not present in the input it was given.

A long-look variant was built for exactly this reason, with 40 frames and a temporal power spectrum per cell type, and it scored 0.826 against 0.871 for the short drift on the same 100 sheets of the predecessor benchmark. That does not settle the question. The same variant lost by the same margin with its spectral features removed, 0.827, because a long drift pushes the window over the padded border and forced pooling over only the central third of the receptor columns. The comparison varied observation length and pooled area together.

Holding the pooled area fixed and varying only the number of analysed frames settles part of it. With the pool frozen, the circuit trails the receptors-only descriptor by $0.011$ at eight frames and leads it by $0.018$, $0.021$ and $0.022$ at twenty, thirty-four and sixty, so the ordering reverses once the observation runs past a single period; the change from eight frames is $+0.033$ $[0.015, 0.050]$. The gain comes from the receptors degrading with length while the circuit holds, which is what a recurrent stage that integrates over time should do. Spectral bands are worth $+0.085$ to $+0.102$ at every length, so the frequency content is real. The mechanism, however, is still not the one we proposed: the T4/T5 contribution does not grow with observation length, at $+0.008$ $[-0.003, 0.018]$, whereas silencing eight random types separates from the intact circuit only at the longest window, by $+0.034$ $[0.022, 0.046]$. These runs use a wider pool and a smaller eye than the headline configuration, so their absolute values are not comparable with the rest of the paper; only the differences within the sweep are.

\subsection{The null we still cannot build}
One null remains out of reach. A wire-length-matched null was attempted and abandoned: only 65{,}577 edges, 1.08 per cent of the graph, have coordinates at both endpoints, and exact length constraints changed eighteen rewiring slots. The topology results therefore establish dependence on the exact wiring given the declared input ports, and do not separate biology from retinotopy and locality. What can be said is narrower than it first appears: exact structure is load-bearing, since every rewiring we can build substantially degrades the result, and it is not sufficient, since the intact circuit still does not beat its own moving retina.

\section{Discussion}
If the data look like the training set, train a small convolutional network; nothing in this study argues otherwise, and the margins are large. If the data are scanned, redrawn at a different stroke weight, or simply produced by a different office, a frozen active-sampling descriptor degrades far more gracefully, and on a closed production set its smallest variant was the better model with two hundred times fewer parameters. The mechanism behind that stability is not biological wiring but the fact that nothing in the pipeline was fitted to the appearance statistics of any particular corpus.

Scientifically the result is smaller, and it has two halves that are easy to confuse. Motion-pretrained parameters transfer to a static document task, and imposed motion contributes measurably on two independent benchmarks. The connectome topology is load-bearing: on both benchmarks every matched rewiring we could build substantially degrades the result, and a positive control shows the test can see a structural change. It is also insufficient: the intact circuit does not beat the moving retina in front of it on either the development or the held-out split. A structure can be load-bearing for a computation without that computation being the one a task needs.

\paragraph{A falsifiable prediction} If the shift result is real and not a single-seed artefact, then in a seeded repetition with cluster bootstrap intervals the accuracy-versus-shift curves of the frozen descriptors should cross those of task-trained networks of every size tested, at a shift no stronger than the scan condition used here. If the curves do not cross, the frozen pipeline has no practical niche in this domain and the contribution is purely scientific.

\paragraph{What the sampling sweep answered} That prediction was run, and it half held. With the pooled area fixed, the circuit does overtake the receptors-only descriptor once the observation passes a single period, by $+0.033$ $[0.015, 0.050]$ relative to the short window, and spectral bands carry a consistent $+0.085$ to $+0.102$. But the mechanism is not the one we named: the T4/T5 contribution does not grow with length. The remaining question is which stage integrates the extra time, and a per-type tuning analysis over the sweep would answer it.

\paragraph{What remains open} The wiring comparison and the single opening of the held-out split resolve the preregistered comparisons, but they do not identify a uniquely biological mechanism. Two gaps remain. The wire-length null cannot be built from the coordinates available, so dependence on exact wiring is not yet separated from dependence on retinotopy and locality. And no real-document benchmark with independent human annotation exists, so every real-sheet number here describes the models under a particular annotation authority rather than under ground truth.

\section{Limitations}
The hatch development split was used repeatedly during design; the held-out split was opened once and reports the same ordering. The moving mask does not guarantee that surrounding context is removed through receptive fields. The frame-order defect, in which the mask was computed on the forward trajectory and not recomputed for reordered sequences, has been repaired by dropping the settling frames before reordering rather than after; drift is now modestly ahead of both, by $+0.007$ $[0.000, 0.013]$ over shuffled frames and $+0.012$ $[0.003, 0.022]$ over reversed ones, so what matters is the presence of motion rather than its direction. Hatch standardization is transductive within a sheet, since it uses the unlabelled queries of that sheet. The receptors-only and network descriptors have been brought to a common dimensionality by fixed random projection; at 512 dimensions the receptors keep the lead in all three seeds, by $+0.032$, $+0.024$ and $+0.034$, so the gap is not an artefact of feature count. The shift matrix now uses three seeds with source-plan intervals. The prescribed motion covers less than one spatial period of the pattern, so no result here should be read as a test of temporal-frequency coding, and the cell-type knockouts inherit that limit. The segmentation experiments use one fixed evaluation crop per source and synthetic geometry, and the full-length convolutional diagnostic was run after the final metrics were disclosed, so it is engineering rather than confirmatory evidence. Real document sets are small and project-imbalanced, one project family may appear under two names across splits, takeoff annotations are partial so unmarked regions are unknown rather than negative, and fill-derived masks omit line-only partitions. No claim about fly perception, behaviour or cognition is made anywhere in this work. No olfactory input, dopaminergic or reward-gated plasticity, and no courtship circuitry exists anywhere in this project; where earlier project material spoke of a material's \emph{smell}, it meant one-shot matching of visual texture and nothing else. Reward functions do exist in other branches of the project, but they are geometric objectives derived from the drawing itself, such as room closure or staying on a wall axis, and no model reported here was trained with one.

\section*{Data, Code and Evidence}
Every number maps, in the accompanying ledger, to a stored run record with its configuration, graph and result hashes; the registry hash chain was verified at audit time and the metric toolkit passed its sixteen tests during manuscript preparation. Model sources are the public \flyvis{} distribution~\cite{turagalab} and the \malecns{} release~\cite{malecns}; plan geometry comes from CubiCasa5K~\cite{kalervo2019} and a public annotated corpus~\cite{cis2025}. The production drawings are proprietary and cannot be released.

An illustrative animation of the active wall-search process is provided as ancillary material.

\paragraph{Use of Generative AI} Generative AI was used in preparing this article.

\clearpage
\onecolumn
\vfill
\begin{center}
\includegraphics[width=0.98\textwidth]{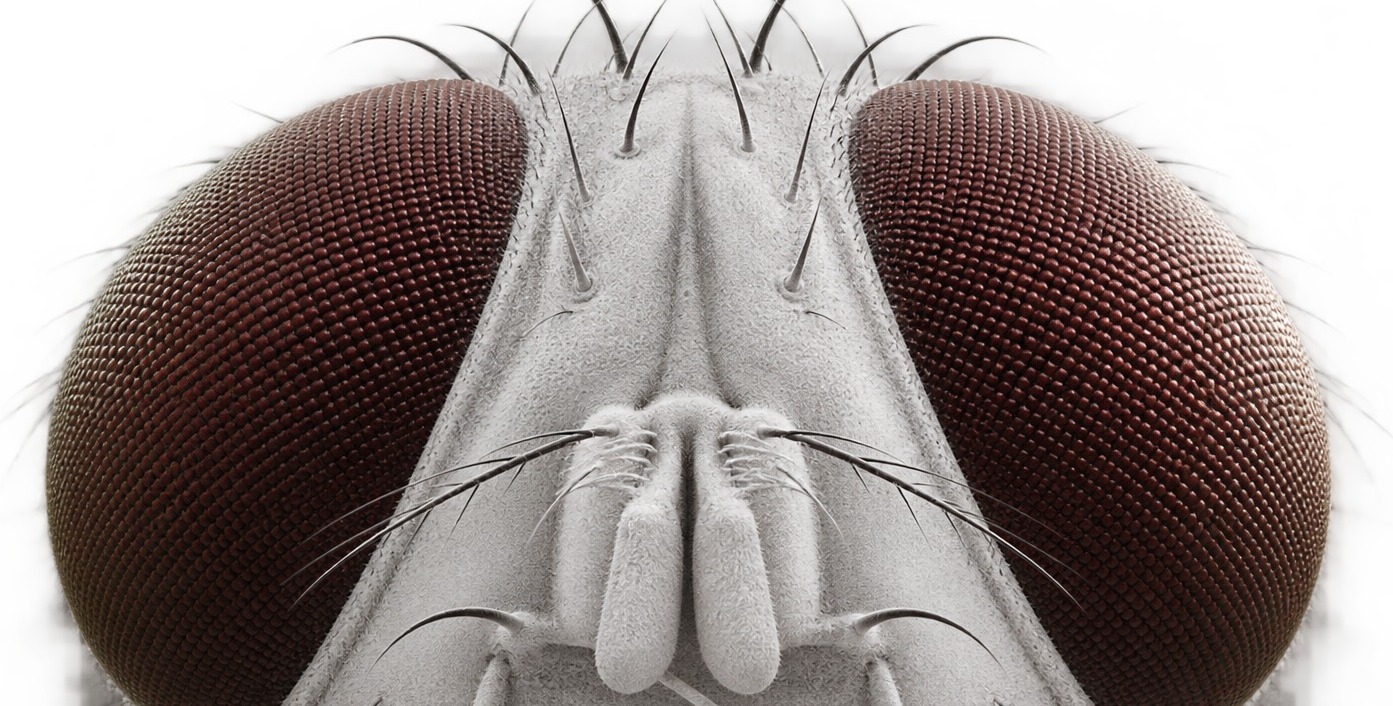}
\end{center}

\end{document}